\documentclass[10pt,twocolumn,letterpaper]{article}

 \usepackage{cvpr}              

\usepackage[dvipsnames]{xcolor}

\definecolor{cvprblue}{rgb}{0.21,0.49,0.74}
\usepackage[pagebackref,breaklinks,colorlinks,allcolors=cvprblue]{hyperref}

\def\paperID{6193} 
\def\confName{CVPR}
\def\confYear{2025}

\usepackage{algorithm}
\usepackage{algpseudocode}
\usepackage{float}

\usepackage{amsthm}

\usepackage{booktabs}
\usepackage{multirow}
\usepackage{array}
\usepackage{graphicx}
\usepackage{xcolor}

\usepackage{array}
\usepackage{multirow}
\usepackage{multicol}
\usepackage{makecell}
\usepackage{tabularx} 
\usepackage{amssymb} 
\usepackage{xcolor} 
\usepackage{pifont} 
\newcommand{\cmark}{\ding{51}} 

\usepackage{caption}
\usepackage{subcaption}
\usepackage{hyperref} 
\usepackage{listings} 
\usepackage{xcolor}   
\usepackage{booktabs}
\usepackage{float}
\usepackage{adjustbox}
\usepackage{placeins} 
\usepackage{colortbl}

\theoremstyle{plain}
\newtheorem*{theorem*}{Theorem}

\title{Q-PART: Quasi-Periodic Adaptive Regression with Test-time Training for Pediatric  Left Ventricular Ejection Fraction Regression}

\author{ Jie~Liu$^1$, Tiexin~Qin$^1$, Hui~Liu$^1$, Yilei~Shi$^{2,3}$, Lichao Mou$^{2,3}$, \\ Xiao Xiang Zhu$^3$, Shiqi Wang$^1$, and Haoliang Li$^1$
\\[2mm]
$^1$City University of Hong Kong~~~~$^2$MedAI Technology (Wuxi) Co. Ltd. \\ $^3$Technische Universität München}

\begin{document}
	\maketitle
	
	\begin{abstract}
		In this work, we address the challenge of adaptive pediatric Left Ventricular Ejection Fraction (LVEF) assessment. While Test-time Training (TTT) approaches show promise for this task, they suffer from two significant limitations. Existing TTT works are primarily designed for classification tasks rather than continuous value regression, and they lack mechanisms to handle the quasi-periodic nature of cardiac signals. To tackle these issues, we propose a novel  \textbf{Q}uasi-\textbf{P}eriodic \textbf{A}daptive \textbf{R}egression with \textbf{T}est-time Training (Q-PART) framework. In the training stage, the proposed Quasi-Period Network decomposes the echocardiogram into periodic and aperiodic components within latent space by combining parameterized helix trajectories with Neural Controlled Differential Equations. During inference, our framework further employs a variance minimization strategy across image augmentations that simulate common quality issues in echocardiogram acquisition, along with differential adaptation rates for periodic and aperiodic components. Theoretical analysis is provided to demonstrate that our variance minimization objective effectively bounds the regression error under mild conditions. Furthermore, extensive experiments across three pediatric age groups demonstrate that Q-PART not only significantly outperforms existing approaches in pediatric LVEF prediction, but also exhibits strong clinical screening capability with high mAUROC scores (up to 0.9747) and maintains gender-fair performance across all metrics, validating its robustness and practical utility in pediatric echocardiography analysis. The project can be found in \href{https://github.com/ljwztc/Q-PART}{Q-PART}.
	\end{abstract}
	
	\section{Introduction}
\begin{figure}[t]
		\centering
		\includegraphics[width=\linewidth]{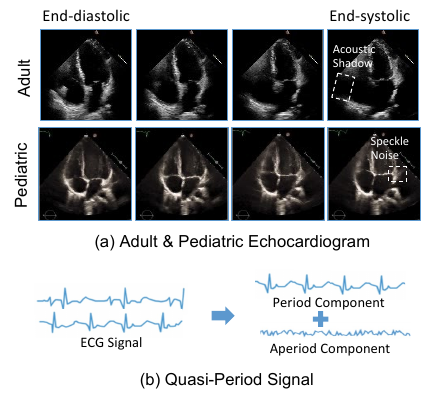}
		\caption{\textbf{(a) Adult and Pediatric Echocardiogram Visualization.} These two kinds of echocardiograms exhibit anatomical and physiological variations, including distinct differences in heart size and shape Moreover, We demonstrate two quality issue, i.e., acoustic shadow and speckle noise. \textbf{(b) Quasi-Period Signal.}  Electrocardiogram (ECG) signals can be decomposed into periodic components originating from regular cardiac cycles and aperiodic components induced by physiological variations and noise.}
		\label{fig:teaser}
	\end{figure}

	Evaluation of Left Ventricular Ejection Fraction (LVEF) for pediatric patients plays a vital role in early diagnosis of congenital and acquired heart disease \cite{reddy2023video}. Echocardiography, recognized for its noninvasive nature and safety, has become the standard modality for LVEF evaluation \cite{ouyang2020video}. Although recent deep learning models \cite{liu2021deep,muhtaseb2022echocotr,li2023echoefnet} have shown promising results in adult LVEF assessment, their direct application to pediatric cases is problematic due to fundamental differences in cardiac imaging. As demonstrated in Figure \ref{fig:teaser} (a), pediatric patients exhibit considerable anatomical and physiological variations, including distinct differences in heart size, shape, and rate across different age groups \cite{eidem2009echocardiography}. Moreover, acquiring pediatric echocardiogram data is inherently challenging. Young children often exhibit difficulty cooperating during examinations, and ethical considerations limit unnecessary medical procedures, resulting in significant inter-observer variability and scarcity of labeled data for training.
	
	Test-time training (TTT) emerges as a promising paradigm to address these challenges by enabling model adaptation to distribution shifts during inference \cite{liang2024comprehensive,schneider2020improving}. TTT can enables model adaptation to each specific patient case during inference, without requiring access to training data or labels \cite{pandey2021generalization,press2024rdumb}, making it particularly valuable for pediatric LVEF assessment. This patient-specific adaptation naturally handles diverse pediatric cases without the necessity for extensive training echocardiograms, while also accommodating diverse anatomical variations and ensuring data privacy.

	However, the regression of pediatric LVEF presents unique challenges for direct application of TTT in this area. \textbf{Firstly}, existing TTT methods such as entropy minimization \cite{wangtent,zhang2022memo,yang2024towards,seto2024realm}, primarily designed for classification tasks with discrete decision boundaries \cite{schneider2020improving,pandey2021generalization,press2024rdumb}, are not well suited for LVEF prediction, which is essentially a continuous value regression task. This challenge is further complicated by the quality issues in echocardiogram acquisition, as shown in Figure \ref{fig:teaser} (a), where varying acoustic shadow, probe positions, and speckle noise significantly impact regression accuracy, necessitating specialized self-supervised algorithms for stable predictions. \textbf{Secondly}, the inherent complexity of cardiac activities in echocardiograms introduces additional challenges. As illustrated by the Electrocardiogram (ECG) signal example in Figure \ref{fig:teaser} (b), cardiac activities demonstrate a characteristic duality, comprising periodic components from regular cardiac cycles and aperiodic components from physiological variations, collectively forming quasi-periodic signals \cite{chakraborty2018efficient,li2023quasi,avendano2023state}. While periodic cardiac patterns remain relatively consistent across age groups, pediatric cases exhibit significant aperiodic variations due to diverse heart sizes and irregular activities. Leveraging this quasi-periodic nature for test-time training in echocardiogram analysis thus presents a promising yet unexplored research direction.

	To address these challenges, we propose a novel test-time training framework, named \textbf{Q}uasi-\textbf{P}eriodic \textbf{A}daptive \textbf{R}egression with \textbf{T}est-time Training (Q-PART), tailored for pediatric LVEF regression with two key innovations: (1) a Quasi-Period Network that explicitly disentangles echocardiogram features into periodic and aperiodic components in latent space, and (2) a specialized variance minimization strategy across augmented samples for continuous value regression. Specifically, the Quasi-Period Network is composed of a latent period encoder that captures fundamental cardiac cycles through a parameterized helix trajectory, and a Neural Controlled Differential Equations (Neural CDE) to represent aperiodic dynamics. This decomposition enables more nuanced feature extraction and adaptation. During test time, we augment test samples with carefully designed augmentation that simulate common quality issues in echocardiogram acquisition, such as acoustic shadows and speckle noise. Then, we introduce a variance minimization objective that minimizes prediction discrepancy across these augmented views, along with a differential adaptation scheme that employs distinct learning rates for periodic and aperiodic components. This design ensures stable regression predictions while effectively handling domain shifts. Furthermore, we provide theoretical guarantees demonstrating that our variance minimization approach effectively bounds the regression error. Extensive experiments demonstrate that our method achieves superior performance compared to existing approaches. Our contributions can be summarized as follows: 
	
	\begin{itemize}
		\item We propose Q-PART, a novel test-time training framework for pediatric LVEF regression. To the best of our knowledge, this \textit{represents the first efforts} to leverage quasi-periodic properties for test-time training in echocardiogram analysis.
		\item We introduce two key technical innovations: a Quasi-Period Network that explicitly disentangles periodic and aperiodic components of echocardiogram, and a variance minimization strategy with theoretical guarantees on regression error bounds.
		\item We conduct comprehensive experiments, demonstrating that Q-PART achieves state-of-the-art performance in pediatric LVEF prediction, exhibits strong clinical screening capability, and maintains gender-fair performance. Additional ablation studies validate the effectiveness of each proposed component.
	\end{itemize}
	
	\section{Related Work}
	\subsection{Automated Echocardiogram Analysis}
	\label{sec:echo_analysis}
	\noindent\textbf{Left Ventricular Segmentation.} Echocardiography is essential for cardiovascular diagnosis and treatment planning \cite{liu2023clip,lai2024echomen,dai2022cyclical}. EchoNet-Dynamic \cite{ouyang2020video} pioneered large-scale echocardiogram video analysis using deep learning framework for left ventricular segmentation in adult patients. This sparked numerous developments in echocardiogram analysis \cite{guo2021dual,wu2022semi,wu2023super}. Through the quantification of left ventricular masks across consecutive frames, the LVEF can be calculated.

	\noindent\textbf{LVEF regression.} Beyond these segmentation-based approaches, several works have explored end-to-end frameworks for direct LVEF prediction \cite{ouyang2020video,dai2022cyclical,lai2024echomen,maani2024coreecho}. Recently, Echo\_CLIP \cite{christensen2024vision} train a large vision-language foundation model for echocardiogram with 1,032,975 cardiac ultrasound videos and corresponding expert text. However, these methods, primarily designed for adult echocardiograms, often struggle with pediatric cases due to significant anatomical variations. 
	
	\noindent\textbf{Quasi-period Property.} Cardiac activities exhibit quasi-periodic characteristics, combining regular cardiac cycles with aperiodic physiological variations. While existing methods have explored quasi-periodic signal processing for one-dimensional signals like ECG \cite{quiroz2017quasiperiodicity,li2023quasi,avendano2023state}, their direct application to high-dimensional spatio-temporal echocardiograms remains challenging. Different from these work, we propose a novel framework that performs quasi-periodic decomposition in the latent space for echocardiogram.

	\subsection{Test-time Training}
	Models often experience significant performance degradation when deployed on test data that differs from the training distribution \cite{panagiotakopoulos2022online}. To address this distribution shift problem, Test-time Training (TTT) has emerged as a promising paradigm that enables models to adapt dynamically during inference \cite{schneider2020improving,pandey2021generalization,press2024rdumb}.
	
	Researchers have employed a self-supervised auxiliary task \cite{hakim2023clust3,sun2020test,gandelsman2022test,liu2021ttt++,osowiechi2023tttflow} to adapt to distribution shifts encountered during test time, from rotation prediction \cite{sun2020test} and masked autoencoding \cite{gandelsman2022test} to contrastive learning \cite{liu2021ttt++} and normalizing flows \cite{osowiechi2023tttflow}. Unlike these general methods, we propose a nuanced domain-specific self-supervised strategy that leverages echocardiogram quality variations through targeted augmentations.
	
	Another mainstream approach uses entropy minimization during inference, pioneered by Tent \cite{wangtent}, which introduced adaptive updates to normalization layers through the mechanism of entropy minimization. This framework has inspired a multitude of subsequent studies \cite{zhang2022memo,yang2024towards,seto2024realm,niu2022efficient,zhao2023delta} that enhance adaptation capabilities via various refinements to the entropy-based objective. Nevertheless, these approaches predominantly target classification tasks, presenting inherent limitations for regression problems, as regression models generate continuous numerical predictions rather than discrete probability distributions required for entropy computation.

	\begin{figure*}[t]
		\centering
		\includegraphics[width=\linewidth]{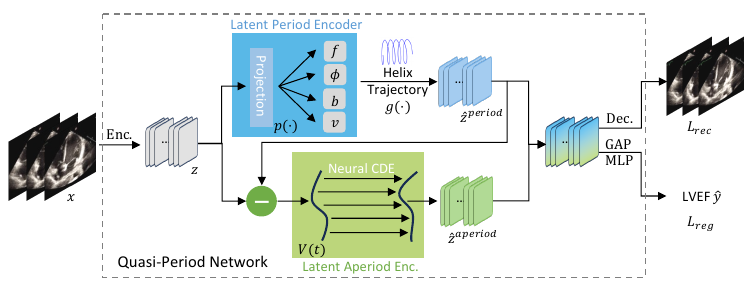}
		\caption{\textbf{Overview of Q-PART Framework.} The Quasi-Period Network decomposes input echocardiogram sequences into periodic and aperiodic components. The periodic component is modeled through a parameterized helix trajectory with frequency $\boldsymbol{f}$, phase shift $\boldsymbol{\phi}$, offset $\boldsymbol{b}$, and velocity $\boldsymbol{v}$. The aperiodic component is captured by Neural CDE. The model is jointly optimized with regression loss $\mathcal{L}_{reg}$ for LVEF prediction and reconstruction loss $\mathcal{L}_{rec}$ for feature learning.}
		\label{fig:framework_train}
	\end{figure*}

	\begin{figure}[t]
		\centering
		\includegraphics[width=\linewidth]{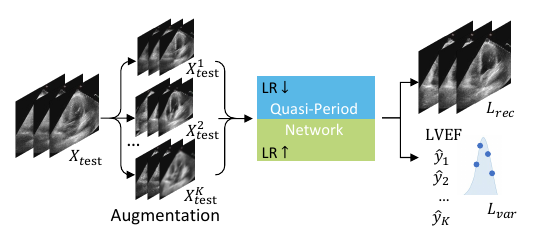}
		\caption{\textbf{~Variance Minimization during Test.} we generate multiple augmented views of the test sequence through domain-specific transformations. The model adapts to each test case by minimizing prediction variance $\mathcal{L}_{var}$ across augmented views and maintaining reconstruction consistency $\mathcal{L}_{rec}$, with different learning rates applied to periodic and aperiodic components.}
		\label{fig:framework_test}
	\end{figure}
	
	\section{Q-PART Framework}
	
	\subsection{Problem Statement}
	Let $\mathcal{D}_{train} = \{(\boldsymbol{x}_i, y_i)\}_{i=1}^N$ denote the training dataset, where $\boldsymbol{x}_i$ represents an echocardiogram sequence with temporal frames and spatial dimensions, and $y_i$ is the corresponding LVEF value. During training, we learn a regression model that maps echocardiogram sequences to LVEF values. The objective of test-time training is to adapt the model for domain-shifted test samples $\boldsymbol{x}_{test} \in \mathcal{P}_{test}$, $\mathcal{P}_{test} \neq \mathcal{P}_{train}$, using self-supervised adaptation objective that does not require ground truth labels. In TTT, the training data are unavailable and the training pipeline cannot be modified.

	\subsection{Overview}
	As illustrated in Figure \ref{fig:framework_train} and Figure \ref{fig:framework_test}, our Q-PART framework consists of a Quasi-Period Network and Variance Minimization during testing time. Given an input sequence $\boldsymbol{x}$, the encoder extracts initial features $\boldsymbol{z}$, which are decomposed into periodic components $\boldsymbol{\hat{z}}^{period}$ through parameterized helix trajectory and aperiodic components $\boldsymbol{\hat{z}}^{aperiod}$ captured by Neural CDEs. During training, the model is optimized with regression loss $\mathcal{L}_{reg}$ and reconstruction loss $\mathcal{L}_{rec}$. At test time, we optimize the model through variance minimization loss $\mathcal{L}_{var}$ and reconstruction loss $\mathcal{L}_{rec}$ using multiple augmented views, with different learning rates for periodic and aperiodic components. Detailed information on important components is provided below.
	
	\subsection{Quasi-Period Network}
	
	Quasi-Period Network addresses the fundamental challenge of disentangling periodic cardiac motion from patient-specific variations in echocardiogram analysis. It decomposes echocardiogram in the latent space through a nuanced architecture with physiologically-informed priors. Notably, during the test phase, \textit{this dual representation allows for adaptive optimization with different learning rates for periodic and aperiodic components, facilitating more effective model adaptation to individual cases.}
	
	\noindent\textbf{Latent Period Encoder.} The latent period encoder architecturally embeds the intrinsic periodicity of cardiac motion into the feature space. Given the initial feature representation $\boldsymbol{z}$, we employ specialized projection head $p$ to compute four fundamental components, frequency $\boldsymbol f$, phase shift $\boldsymbol \phi$, static offset $\boldsymbol b$, and velocity $\boldsymbol v$. This parameterization process can be formalized as
	\begin{equation}
		\boldsymbol f,\boldsymbol \phi,\boldsymbol b,\boldsymbol v = p(\boldsymbol{z}),
	\end{equation}
	where $\boldsymbol f,\boldsymbol \phi,\boldsymbol b,\boldsymbol v \in \mathbb{R}^{c\times h \times w}$ represents the global periodic characteristics for sequence $\boldsymbol{x}$, maintaining $c$ latent channels and spatial dimensions $h \times w$ to preserve spatial structures. To explicitly represent the periodicity, we define a parameterized helix trajectory function $g(\cdot)$ in multi-dimensional space, motivated by the physiological nature of cardiac motion that exhibits spiral-like patterns through combined rotational and translational movements:
	\begin{equation}
		\begin{split}
			\boldsymbol{\hat{z}}^{period}_t = g(t)&= \cos(2\pi (\boldsymbol f t- \boldsymbol \phi)) \\
			& + \sin(2\pi (\boldsymbol f t - \boldsymbol \phi)) + \boldsymbol v t + \boldsymbol b.
		\end{split}
	\end{equation}
	where $\boldsymbol{\hat{z}}^{period}_t \in \mathbb{R}^{c \times h \times w}$ represents the period component in time $t$.
	The sinusoidal terms capture the intrinsic periodicity of cardiac motion, while the linear term $\boldsymbol v t + \boldsymbol b$ accounts for long-term translational movements. This formulation creates a helix trajectory that naturally integrates both cyclic cardiac motion and progressive physiological dynamics.

	\noindent\textbf{Latent Aperiodic Encoder.} To capture the inherent non-periodic variations in cardiac motion, we design a latent aperiodic encoder that explicitly models these irregular components. The aperiodic features are obtained as the residual between the initial representation and the periodic component: $\boldsymbol{z}' = \boldsymbol z - \boldsymbol{\hat{z}}^{period}$. While traditional RNNs could potentially model temporal variations, Neural CDEs offer these crucial advantages for our task: 1) they naturally handle irregular representation in echocardiogram sequences, (2) they learn continuous-time dynamics from discretely sampled observations, which is better aligned with physiological processes.
	
	Formally, given the aperiodic component $\boldsymbol{z}'$, we first construct a continuous path $\boldsymbol{V}(t)$ through natural cubic spline \cite{morrill2022choice} to controlled differential equations, $\boldsymbol{V}(t) = \text{CubicSpline}(\boldsymbol{z}'_t)$, where $\boldsymbol{z}'_t$ represents the aperiodic component $\boldsymbol{z}'$ at time $t$. The spline ensures smooth transitions between discrete observations while preserving the underlying temporal structure. The evolution of the aperiodic representation is then governed by the controlled differential equation, $d\boldsymbol{\hat{z}}^{aperiod}_t = f_{\theta}(\boldsymbol{\hat{z}}^{aperiod}_t, t) d\boldsymbol{V}(t)$, where $f_{\theta}$ is implemented as a neural network that learns the mapping between the current state and its differential elements. The final aperiodic representation at time $T$ is obtained through integration,
	\begin{equation}
		\boldsymbol{\hat{z}}^{aperiod}_T = \boldsymbol{z}'_0 + \int_0^T f_{\theta}(\boldsymbol{\hat{z}}^{aperiod}_t, t) d\boldsymbol{V}(t),
	\end{equation}
	where $\boldsymbol{z}'_0$ serves as the initial condition. The neural CDE network is optimized using the CDE solver \cite{kidger2020neural}, which compute the gradients via the adjoint method. 
	
	The decomposition provides a foundation for our test-time training strategy, where periodic and aperiodic components can be optimized with different learning dynamics. Empirically, we find that this architecture achieves superior performance in capturing complex cardiac motions, as demonstrated in our ablation studies \S \ref{sec:ablation}.
	
	\subsection{Variance Minimization}
	To address the domain shift challenge during testing, we propose a theoretically-grounded self-supervised strategy that combines test-time augmentation with variance minimization. Our key insight is that while echocardiographic acquisitions may vary in quality and appearance, the underlying cardiac function remains consistent across different views of the same sequence.
	
	\noindent\textbf{Domain-Specific Augmentation.} For each test sequence $\boldsymbol{x}_{test}$, we generate $K$ augmented versions through a carefully curated augmentation pipeline:
	\begin{equation}
		\{\boldsymbol{x}_{test}^k\}_{k=1}^K = \mathcal{T}_{aug}(\boldsymbol{x}_{test}),
	\end{equation}
	where $\mathcal{T}_{aug}$ comprises three categories of domain-specific augmentations. Beyond conventional image augmentations), we introduce three ultrasound-specific transformations: gamma correction to simulate varying image contrast settings, speckle noise injection to mimic acoustic interference patterns, and elastic deformation to account for cardiac tissue deformation during acquisition. These domain-specific augmentations are specifically designed to simulate real-world variations in echocardiographic acquisition while preserving the underlying cardiac motion patterns essential for LVEF assessment.
	
	\noindent\textbf{Variance-Based Adaptation.} Then, we minimizes the prediction variance across augmented samples. The variance loss $\mathcal{L}_{var}$ is computed as:
	\begin{equation}
		\mathcal{L}_{var} = \frac{1}{K}\sum_{k=1}^K(\hat{y}_k - \bar{y})^2,
		\label{eq:variance}
	\end{equation}
	where $\hat{y}_k$ represents the prediction for the $k$-th augmented sequence and $\bar{y}$ denotes the mean prediction across all augmentations. This formulation encourages the model to learn invariant features that are robust to acquisition-specific variations while remaining sensitive to true cardiac function changes. We provide theoretical analysis in Section \ref{sec:theoretical_analysis} demonstrating that minimizing this variance effectively bounds the adaptation error under mild conditions.
	
	\subsection{Optimization}
	
	Our optimization strategy consists of two phases: training time and testing time. The overall pipeline is summarized in Appendix Algorithm \ref{alg:ttt_qpn}.
	
	\noindent \textbf{Training time.} During the training phase, our model employs a multi-objective learning strategy combining regression and reconstruction objectives:
	\begin{equation}
		\mathcal{L}_{total} = \mathcal{L}_{reg} + \mathcal{L}_{rec}.
		\label{eq:train_total}
	\end{equation}
	The reconstruction loss ensures that both periodic and aperiodic components collectively preserve the essential information of the input sequence.
	\begin{equation}
		\mathcal{L}_{rec} = \|\boldsymbol{x} - \text{Dec}(\boldsymbol{\hat{z}}^{period} + \boldsymbol{\hat{z}}^{aperiod})\|_2^2
	\end{equation}
	where Dec($\cdot$) denotes the decoder network. For LVEF prediction, we aggregate the combined latent representation through Global Average Pooling (GAP) followed by a Multiple Layer Perceptron (MLP), $\hat{y} = \text{MLP}(\text{GAP}(\boldsymbol{\hat{z}}^{period} + \boldsymbol{\hat{z}}^{aperiod}))$. The regression loss is defined as:
	\begin{equation}
		\mathcal{L}_{reg} = \|\hat{y} - y\|_2^2,
	\end{equation}
	Through this joint optimization, our model learns to decompose the echocardiogram into meaningful periodic and aperiodic representation while maintaining accurate LVEF predictions.
	
	\noindent \textbf{Testing Time.} The overall test optimization objective combines this variance minimization with reconstruction consistency
	\begin{equation}
		\mathcal{L}_{test} = \mathcal{L}_{var} + \mathcal{L}_{rec}.
		\label{eq:loss_test}
	\end{equation}
	During testing time, we only update the batch normalization (BN) parameters in our network \cite{sun2020test}, which has been shown to be effective for test-time training while maintaining model stability. Given the inherent differences between periodic and aperiodic cardiac components, we employ different learning rates for their respective BN parameters. The periodic components, which capture the fundamental cardiac cycles, require more conservative updates to maintain temporal consistency. Therefore, we use a smaller learning rate for BN layers in the latent period encoder. In contrast, the aperiodic components, representing physiological variations and acquisition-specific features, benefit from more aggressive adaptation.

	\subsection{Theoretical Analysis}
	\label{sec:theoretical_analysis}
	To theoretically justify the effectiveness of our test-time training strategy, we establish the relationship between prediction variance and regression error. Our theoretical framework demonstrates how variance minimization across augmented samples inherently reduces the model's prediction error.
	
	\textbf{Assumption 1} \textit{(Unbiased Augmentation) The expected prediction over augmentations equals the ground truth LVEF: $\mathbb{E}_{\mathcal{T}_{aug}}[\hat{y}_k] = y$.}
	
	This assumption is grounded in the design of our augmentation strategy $\mathcal{T}_{aug}$, which simulates acquisition-specific variations (e.g., speckle noise, gamma correction) while preserving the underlying cardiac motion patterns that determine LVEF. These perturbations affect image quality but do not alter the fundamental cardiac functional measurements.
	
	\textbf{Assumption 2} \textit{(Augmentation Independence) The predictions from different augmented samples are independently distributed, with zero cross-covariance: $\operatorname{Cov}_{\mathcal{T}_{aug}}(\hat{y}_k, \hat{y}_j) = 0$ for $k \neq j$.}
	
	This assumption reflects our augmentation design principle where each transformation introduces independent variations. The independence is achieved through combining multiple orthogonal perturbation types, ensuring minimal correlation between augmented samples.
	
	\textbf{Theorem 1} \textit{Consider a test sample $\boldsymbol{x}_{test}$ with ground truth label $y$, and its $K$ augmented versions $\{\boldsymbol{x}_{test}^k\}_{k=1}^K$ generated by $\mathcal{T}_{aug}$. Under Assumptions 1 and 2, the expected regression loss $\mathbb{E}[L_{reg}] = \mathbb{E}[(\hat{y} - y)^2]$ is upper bounded by:
		\begin{equation}
			\mathbb{E}[L_{reg}] \leq \frac{2\mathbb{E}[L_{var}]}{K},
		\end{equation}
		where $L_{var}$ is the variance loss defined in Equation \ref{eq:variance}.}
	
	The proof uses standard techniques in optimization, and is left for the \autoref{append_sec_proof}. This theoretical result provides a justification for our test-time training strategy. Minimizing the variance loss directly leads to a reduction in the expected regression error, with the bound becoming tighter as we increase the number of augmentations $K$.

	\section{Experiment}
	\subsection{Experiment Setup}
	\noindent\textbf{Dataset and Metrics.} We evaluate Q-PART using two benchmark datasets of apical-4-chamber echocardiogram videos collected from EchoNet-Dynamic \cite{ouyang2019echonet} and EchoNet-Pediatric \cite{reddy2023video}. Following clinical standards, we stratify the data into four age cohorts: pre-school (ages 0-5), school-age (ages 6-11), adolescence (ages 12-18), and adult (ages $>$ 18). The detailed statistics of each cohort are summarized in Table \ref{tab:dataset_statistics}. We train the model on the adult cohort and assess its performance through test-time training on pre-school, school-age, and adolescence cohorts respectively. For evaluation metrics, we adopt three widely-used metrics to comprehensively evaluate the LVEF prediction performance: Mean Absolute Error (MAE), Root Mean Square Error (RMSE), and Mean Absolute Percentage Error (MAPE). Moreover, we evaluate the model's clinical screening capability through mean Area Under the Receiver Operating Characteristic curve (mAUROC) analysis at multiple LVEF thresholds (35\%, 40\%, 45\%, and 50\%). More details please refer to Appendix \ref{sec:append_metrics}.
	
	\begin{table}[t]
	\caption{\textbf{Dataset Statistics.} Demographic and clinical characteristics across age groups. The statistics show substantial differences in sample sizes (ranging from 831 to 7465), with notably fewer samples in younger age groups due to the inherent challenges in pediatric data collection.}
	\label{tab:dataset_statistics}
	\centering
	\resizebox{\linewidth}{!}{
		\begin{tabular}{lccccc}
			\toprule
			Statistics & Pre-school & School-age & Adolescence & Adult \\
			\midrule
			\#Samples & 831 & 914 & 1, 539 & 7, 465 \\
			Age & 2 ± 1 & 9 ± 2 & 15 ± 2 & 70 ± 22 \\
			Female (\%) & 404 (49\%) & 371 (41\%) & 617 (40\%) & 3662 (49\%) \\
			LVEF (\%) & 59.6 ± 12.8 & 62.0 ± 9.2 & 61.0 ± 9.8 & 55.8 ± 12.4 \\
			\bottomrule
		\end{tabular}
	}
\end{table}

	\noindent\textbf{Implementation.} We adopt R3D Network \cite{tran2018closer} as the backbone encoder. The projection head contains four groups, with each group consisting of two 1×1 convolutional layers (512 → 64 → 64) with batch normalization and ReLU activation. The decoder uses five transposed convolutional layers for reconstruction. In training stage, The model is optimized using SGD with momentum 0.9 and weight decay 1e-4, with a linear warmup for the first 10 epochs followed by cosine annealing for the remaining 35 epochs. During test-time training, we employ both general augmentations (rotation, flips, grid distortion, Gaussian blur) and ultrasound-specific transformations (speckle noise, gamma correction, acoustic shadow). We apply different learning rates to periodic (1e-5) and aperiodic (1e-3) components, while maintaining a base learning rate of 1e-4 for other parameters. The value K is 8. All experiments are conducted on a single NVIDIA A100 80GB GPU.

	\begin{table*}[t]
	\centering
	\setlength{\tabcolsep}{4pt}  
	\newcolumntype{P}[1]{>{\centering\arraybackslash}p{#1}} 
	
	\caption{\textbf{Comparison with State-of-the-Art Methods.} We benchmark Q-PART against representative approaches spanning three methodological categories: segmentation-based methods (Seg-Based), vision-language models (VLM), and end-to-end regression approaches (E2E). Our method demonstrates superior adaptation capability and robustness across age groups, particularly in challenging scenarios with younger patients. All results are reported with four significant digits. \textbf{Bold} numbers indicate the best performance.}
	\resizebox{\linewidth}{!}{
		\begin{tabular}{ll|P{1.5cm}P{1.5cm}P{1.5cm}|P{1.5cm}P{1.5cm}P{1.5cm}|P{1.5cm}P{1.5cm}P{1.5cm}}
			\toprule
			Type & Approach & \multicolumn{3}{c|}{Pre-School} & \multicolumn{3}{c|}{School Age} & \multicolumn{3}{c}{Adolescence} \\
			& & MAE & MAPE & RMSE & MAE & MAPE & RMSE & MAE & MAPE & RMSE \\
			\midrule
			\multirow{2}{*}{Seg-Based} 
			& EchoNet \cite{ouyang2020video} {\footnotesize\textcolor{gray}{(Nature 2020)}} 
			& 20.90 & 0.4730 & 26.14 
			& 17.86 & 0.3403 & 21.97 
			& 19.64 & 0.3761 & 23.90 \\
			& MemSAM \cite{deng2024memsam} {\footnotesize\textcolor{gray}{(CVPR 2024)}} 
			& 17.94 & 0.3675 & 21.97 
			& 18.35 & 0.3216 & 21.76 
			& 18.71 & 0.3249 & 22.26 \\
			\midrule
			VLM 
			& EchoCLIP \cite{christensen2024vision} {\footnotesize\textcolor{gray}{(Nature Med. 2024)}} 
			& 11.56 & 0.2035 & 14.18 
			& 9.303 & 0.1604 & 11.71 
			& 8.496 & 0.1521 & 10.93 \\
			\midrule
			\multirow{5}{*}{E2E} 
			& Source-only 
			& 9.114 & 0.1759 & 10.86 
			& 8.425 & 0.1469 & 10.03 
			& 8.636 & 0.1549 & 10.53 \\
			& TTT++ \cite{liu2021ttt++} {\footnotesize\textcolor{gray}{(NeurIPS 2022)}} 
			& 8.053 & 0.1669 & 10.01 
			& 7.430 & 0.1316 & 9.096 
			& 7.616 & 0.1416 & 9.601 \\
			& Clust3 \cite{hakim2023clust3} {\footnotesize\textcolor{gray}{(CVPR 2023)}} 
			& 9.292 & 0.1852 & 11.23 
			& 7.892 & 0.1393 & 9.424 
			& 8.216 & 0.1509 & 9.995 \\
			& NC-TTT \cite{osowiechi2024nc} {\footnotesize\textcolor{gray}{(CVPR 2024)}} 
			& 8.352 & 0.1800 & 10.45 
			& 7.285 & 0.1325 & 8.962 
			& 7.422 & 0.1408 & 9.347 \\
			\cmidrule(lr){2-11}
			\rowcolor[gray]{0.8} & Q-PART {\footnotesize\textcolor{gray}{(Ours)}} 
			& \textbf{7.235} & \textbf{0.1611} & \textbf{9.290} 
			& \textbf{6.706} & \textbf{0.1244} & \textbf{8.432}
			& \textbf{6.980} & \textbf{0.1344} & \textbf{8.950} \\
			\bottomrule
		\end{tabular}
	}
	\label{tab:state_of_the_art}
\end{table*}
	
	\begin{figure*}[t]
		\centering
		\includegraphics[width=\linewidth]{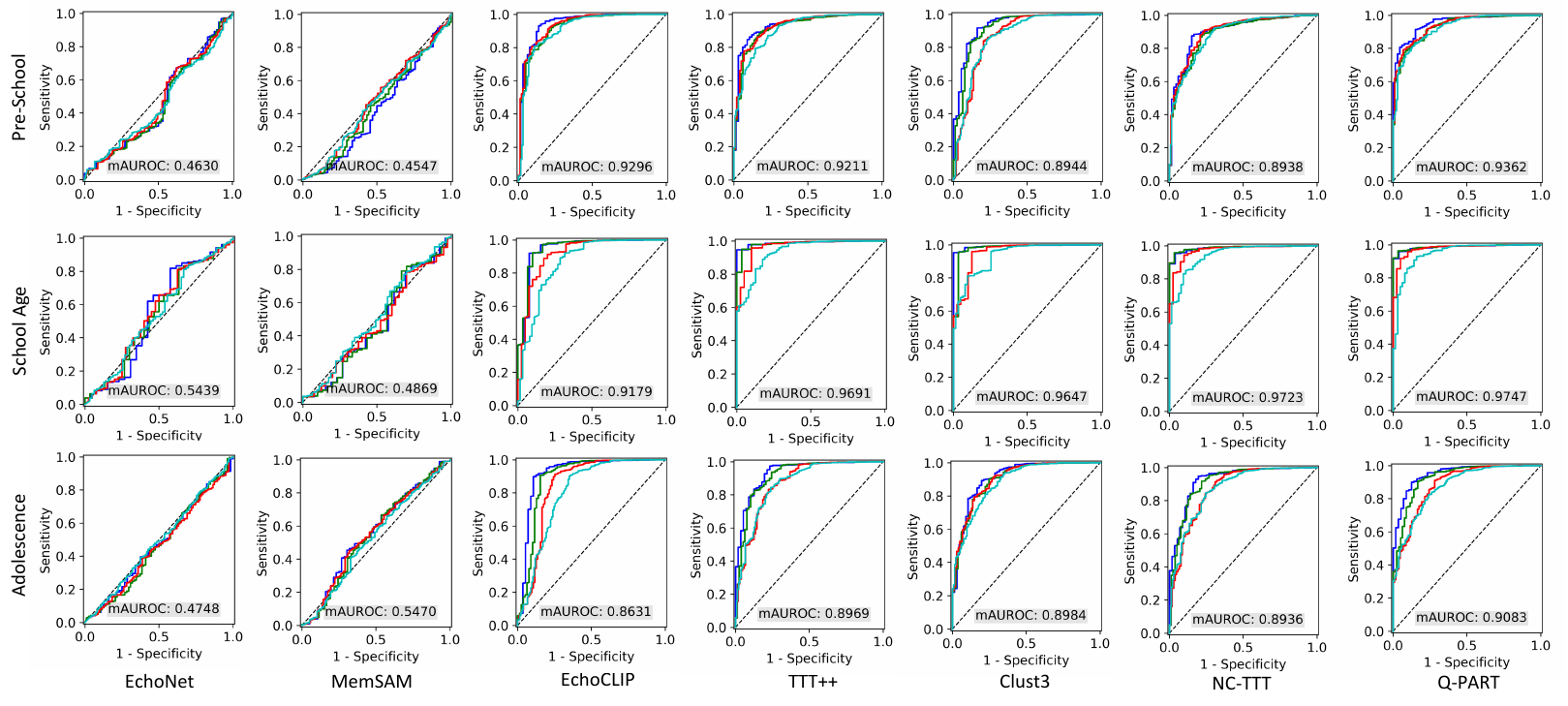}
		\caption{\textbf{ROC Analysis at Multiple Clinical Thresholds. Please zoom in for better view.} Receiver Operating Characteristic (ROC) curves for LVEF regression with four clinically significant thresholds: 35\% (\textcolor{blue}{blue}), 40\% (\textcolor{green}{green}), 45\% (\textcolor{red}{red}), and 50\% (\textcolor{cyan}{cyan}). These thresholds represent different degrees of cardiac dysfunction severity. The x-axis shows 1-specificity (false positive rate) and the y-axis shows sensitivity (true positive rate). The mAUROC is shown in \colorbox{gray!20}{gray box}. A high mAUROC value indicates strong clinical reliability. Higher sensitivity ensures early detection of cardiac dysfunction, while higher specificity reduces unnecessary follow-up examinations and patient anxiety, collectively supporting more accurate clinical decision-making and resource allocation.}
		\label{fig:exp_roc}
	\end{figure*}
	
	\subsection{Comparison with State-of-the-art Methods}
	We comprehensively evaluate Q-PART against seven representative methods spanning three methodological categories, i.e., segmentation-based approaches, vision-language models, and end-to-end regression methods, as mentioned in \S \ref{sec:echo_analysis}. Three key findings emerge from our experiments:
	
	\noindent\textbf{1) Superior Performance.} As shown in Table \ref{tab:state_of_the_art}, our method achieves consistent improvements across all metrics and age groups, with particularly notable gains in the challenging Pre-School cohort (MAE: 7.235, MAPE: 0.1611, RMSE: 9.290). This demonstrates Q-PART's robust adaptation capability and its effectiveness in handling the unique challenges of pediatric echocardiography.
	
	\noindent\textbf{2) Clinical Screening Capability.} Beyond regression metrics, we evaluate our model's clinical utility through ROC analysis at multiple LVEF thresholds in Figure \ref{fig:exp_roc}. The high mAUROC scores have significant clinical implications: the enhanced sensitivity ensures reliable detection of cardiac dysfunction cases requiring immediate intervention, while the improved specificity reduces unnecessary follow-up examinations, thereby optimizing healthcare resource allocation and minimizing patient anxiety. Q-PART achieves remarkable mAUROC scores of 0.9362, 0.9747, and 0.9083 for Pre-School, School Age, and Adolescence cohorts, respectively, substantially outperforming other TTA methods. This superior performance validates the effectiveness of our echocardiogram-specific framework. 
	
	\noindent\textbf{3) Segmentation vs End-to-End Approaches.} As discussed in \S\ref{sec:echo_analysis}, while lots of methods rely on left ventricular segmentation to derive LVEF values, recent end-to-end approaches directly predict LVEF. Our experimental results in Table \ref{tab:state_of_the_art} strongly support the superiority of end-to-end approaches. Despite advances in segmentation accuracy with methods like EchoNet and MemSAM, their LVEF estimation performance (MAE $>$ 17.94) lags significantly behind end-to-end approaches (MAE $<$ 9.114). This substantial performance gap suggests that accurate segmentation does not necessarily translate to precise LVEF estimation, as the latter requires capturing complex temporal dynamics and subtle motion patterns that may be lost in the intermediate segmentation step.

	\begin{table}[t]
	\centering
	\setlength{\tabcolsep}{4pt}  
	\newcolumntype{P}[1]{>{\centering\arraybackslash}p{#1}} 
	
	\caption{\textbf{Ablation Study of Key Components.} Analysis of three key components: QP-Net (Quasi-Period Network), LR (Learning Rate Strategy), and VM (Variance Minimization). We presents the results on Pre-School cohort here, which demonstrates the effectiveness of each component. Complete results across all age cohorts are presented in Appendix Table \ref{tab:ablation_full}.}
	\resizebox{\linewidth}{!}{
		\begin{tabular}{P{1.1cm}P{1.1cm}P{1.1cm}|P{1.2cm}P{1.2cm}P{1.2cm}}
			\toprule
			QP-Net &  LR & VM &  MAE & MAPE & RMSE\\
			\midrule
				& & 
			& 8.129 & 0.1681 & 10.09\\
			\rowcolor[gray]{0.9} \cmark & & 
			& 7.949 & 0.1666 & 9.909 \\
			\cmark & \cmark & 
			& 7.909 & 0.1656 & 9.878 \\
			\rowcolor[gray]{0.9} & & \cmark
			& 7.842 & 0.1640 & 9.714 \\
			\cmark&  & \cmark
			& 7.283 & 0.1619 & 9.307 \\
			\rowcolor[gray]{0.9} \cmark&\cmark &\cmark 
			& 7.235 & 0.1611 & 9.290 \\
			\bottomrule
		\end{tabular}
	}
	\label{tab:ablation}
\end{table}

	\subsection{Ablation Study}
	\label{sec:ablation}
	We conduct comprehensive ablation studies to evaluate the contribution of each key component in Q-PART, i.e., Quasi-Period Network (QP-Net), Learning Rate Strategy (LR), and Variance Minimization (VM). Table \ref{tab:ablation} presents the results on the Pre-School cohort. Several observations can be drawn from the ablation results: (1) The Quasi-Period Network alone yields substantial improvements (MAE: 7.949), demonstrating the effectiveness of disentangling periodic and aperiodic components during test-time training. (2) When combining QP-Net with VM, we observe a notable performance gain (MAE: 7.283), validating our hypothesis that variance minimization across augmented views effectively guides model adaptation. (3) The differential learning rate strategy further refines the adaptation process, with our full model achieving the best performance (MAE: 7.235). This suggests that applying different learning rates to periodic and aperiodic components helps maintain stable cardiac cycle patterns while allowing flexible adaptation to patient-specific variations.
	
	\begin{figure}[t]
		\centering
		\includegraphics[width=0.85\linewidth]{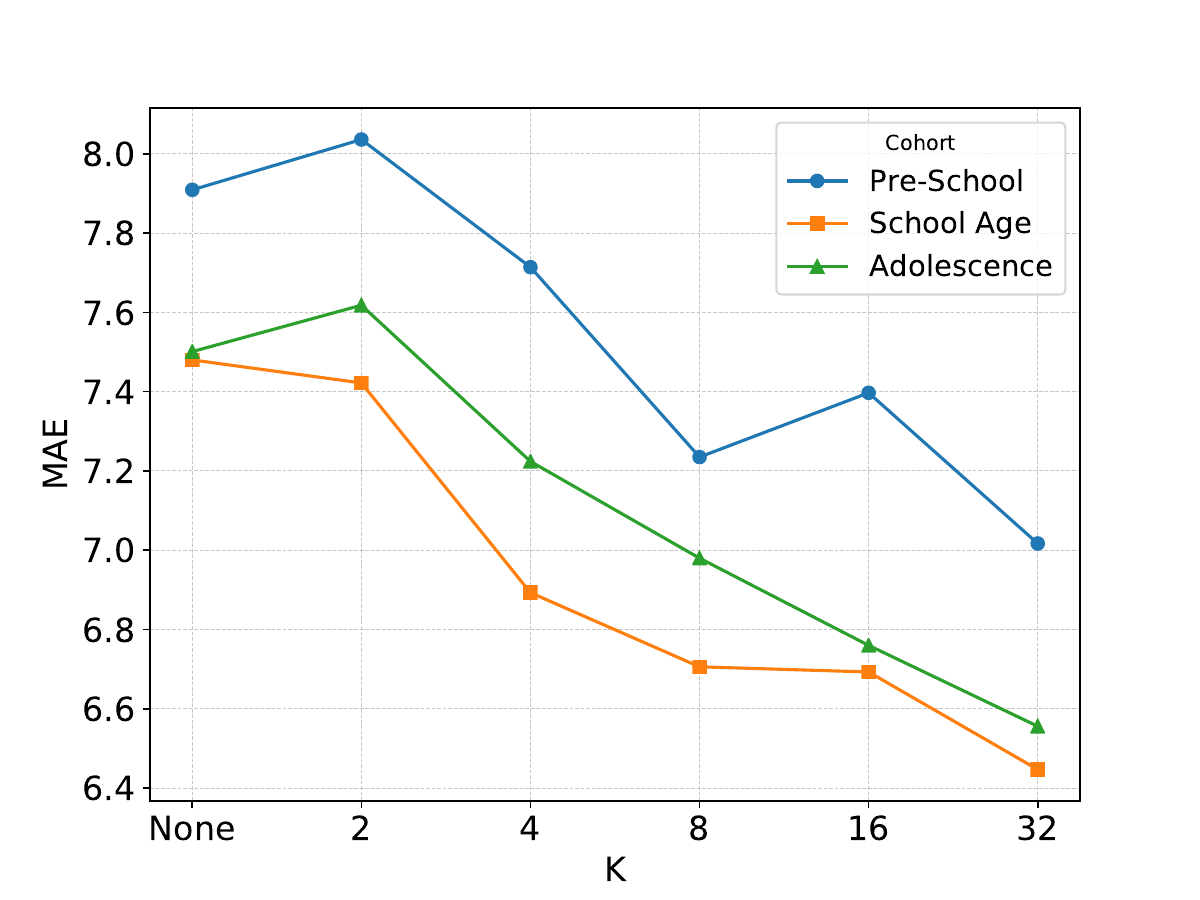}
		\caption{\textbf{Impact of $K$ Augmentation Number on Model Performance.} The plot shows how performance varies with different numbers of augmentation number across three age cohorts.}
		\label{fig:number_k}
	\end{figure}
	
	\begin{figure}[t]
		\centering
		\includegraphics[width=\linewidth]{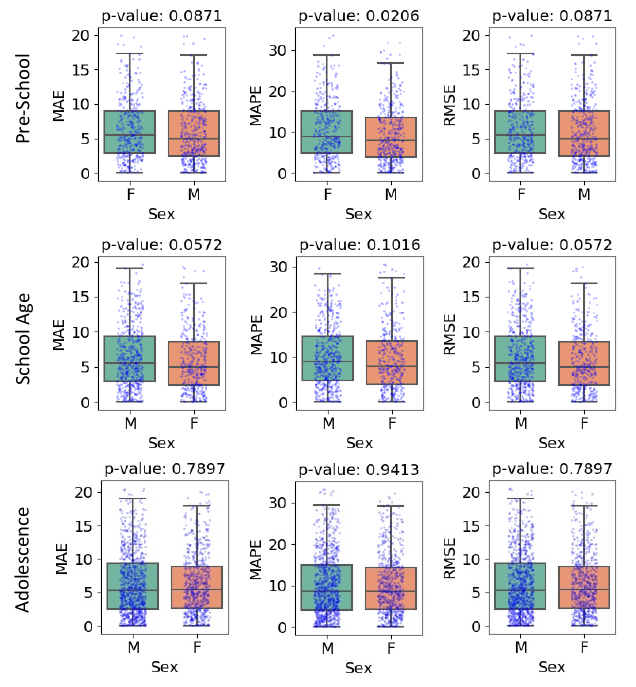}
		\caption{\textbf{Gender Bias Analysis Across Age Cohorts.} Box plots comparing model performance between male and female subjects across three age cohorts using three metrics. Each subplot shows the distribution of performance metrics for both genders, with p-values from Mann-Whitney U tests. The absence of statistically significant differences (p $>$ 0.05) in eight out of nine comparisons demonstrates the model's gender-fair adaptation capability.}
		\label{fig:exp_sex}
	\end{figure}

	\subsection{Analysis}
	\noindent\textbf{Impact of Augmentation Number $K$.} We investigate how the number of augmented views $K$ in variance minimization affects model performance. As shown in Figure \ref{fig:number_k}, larger $K$ values generally lead to better performance across all age cohorts. Notably, when $K=2$, the model performs worse than the baseline without variance minimization, with MAE increasing from 7.909 to 8.036 in the Pre-School cohort. This deterioration can be attributed to the insufficient number of samples for reliable variance estimation, resulting in unstable optimization signals during test-time adaptation. While further increasing $K$ to 32 continues to yield marginal improvements, it significantly increases computational overhead during inference. Balancing performance gains against computational efficiency, we choose $K=8$ as our default setting for all main experiments.

	\noindent\textbf{Gender-fair Performance.} To assess potential gender bias, we conduct a comprehensive analysis comparing model performance between male and female subjects across all age cohorts and metrics, as visualized in Figure \ref{fig:exp_sex}. Statistical analysis reveals no significant differences (p $>$ 0.05 in eight out of nine comparisons) between gender groups across all evaluation metrics (MAE, MAPE, RMSE) and age cohorts. This gender-fair performance demonstrates that Q-PART maintains consistent reliability regardless of gender, which is crucial for ensuring equitable healthcare delivery and maintaining diagnostic consistency across diverse patient populations.

	\section{Conclusion}
	We present Q-PART, a novel test-time training framework specifically designed for pediatric LVEF regression. By explicitly disentangling periodic cardiac motion from patient-specific variations through our Quasi-Period Network, and employing a theoretically-grounded variance minimization strategy, Q-PART effectively addresses the unique challenges in pediatric echocardiography. Extensive experiments demonstrate that our method significantly outperforms existing approaches across different age groups and evaluation metrics. Q-PART advances the field of automated cardiac function assessment and shows promise for improving clinical workflow in pediatric cardiology.
	
	{
		\small
		\bibliographystyle{ieeenat_fullname}
		\bibliography{main}
	}
	
	\clearpage
	\appendix
	\twocolumn[{
		\begin{@twocolumnfalse}
			\maketitle
			\begin{center}
				\textbf{\LARGE{Appendix for Q-PART}}
			\end{center}
		\end{@twocolumnfalse}
	}]
	
	\noindent\textbf{Abstract.} In this appendix, we provides supplementary materials and detailed explanations for the main paper. \autoref{append_sec_proof} presents a comprehensive mathematical proof of Theorem 1. \autoref{appenx_sec_algo} elaborates on our algorithm implementation, providing detailed descriptions of both training and test-time adaptation phases of the proposed Q-PART framework. \autoref{append_sec_exp} contains additional experimental details, including formal definitions of evaluation metrics, detailed implementation of baseline methods, and complete ablation study results across all age cohorts.

	\section{Proof of Theorem 1}
	\label{append_sec_proof}
	\textbf{Proof:}
	We begin by computing the expected variance loss \(\mathbb{E}[L_{\text{var}}]\).
	The variance loss is defined as
	\begin{equation}
		L_{\text{var}} = \frac{1}{K}\sum_{k=1}^K (\hat{y}_k - \bar{y})^2.
	\end{equation}
	Taking the expected value
	\begin{equation}
		\mathbb{E}[L_{\text{var}}] = \frac{1}{K} \sum_{k=1}^K \mathbb{E}\left[(\hat{y}_k - \bar{y})^2\right].
	\end{equation}
	Since \(\mathbb{E}\left[\hat{y}_k - \bar{y}\right] = \mathbb{E}[\hat{y}_k] - \mathbb{E}[ \bar{y}] = 0\), each term \(\mathbb{E}\left[(\hat{y}_k - \bar{y})^2\right]\) can be expressed using the variance formula:
	\begin{equation}
		\mathbb{E}\left[(\hat{y}_k - \bar{y})^2\right] = \operatorname{Var}(\hat{y}_k - \bar{y}).
	\end{equation}
	Then, noted that:
	\begin{equation}
		\operatorname{Var}(\hat{y}_k - \bar{y}) = \operatorname{Var}(\hat{y}_k) + \operatorname{Var}(\bar{y}) - 2\,\operatorname{Cov}(\hat{y}_k, \bar{y}).
	\end{equation}
	Set the variance of the predictions from the augmented samples $\hat{y}_k$ as $\sigma^2$,
	\begin{equation}
		\operatorname{Var}(\hat{y}_k) = \sigma^2.
	\end{equation}
	Under Assumption 2 (Augmentation Independence), the predictions \(\hat{y}_k\) are independent for different \(k\). Compute the $\operatorname{Var}(\bar{y})$
	\begin{equation}
		\begin{split}
			\operatorname{Var}(\bar{y}) = \operatorname{Var}\left(\frac{1}{K}\sum_{k=1}^K \hat{y}_k\right) &= \frac{1}{K^2} \sum_{k=1}^K \operatorname{Var}(\hat{y}_k) \\
			&= \frac{1}{K^2} \cdot K \cdot \sigma^2 = \frac{\sigma^2}{K}.
		\end{split}
	\end{equation}
	Now, compute the covariance \(\operatorname{Cov}(\hat{y}_k, \bar{y})\):
	\begin{equation}
		\begin{split}
			\operatorname{Cov}(\hat{y}_k, \bar{y}) &= \operatorname{Cov}\left(\hat{y}_k, \frac{1}{K}\sum_{j=1}^K \hat{y}_j\right) \\
			&= \frac{1}{K} \operatorname{Cov}(\hat{y}_k, \hat{y}_k) + \frac{1}{K} \sum_{j \neq k} \operatorname{Cov}(\hat{y}_k, \hat{y}_j).
		\end{split}
	\end{equation}
	Since \(\operatorname{Cov}(\hat{y}_k, \hat{y}_k) = \operatorname{Var}(\hat{y}_k) = \sigma^2\), and under Assumption 2, for \(j \neq k\), \(\operatorname{Cov}(\hat{y}_k, \hat{y}_j) = 0\), we have:
	\begin{equation}
		\operatorname{Cov}(\hat{y}_k, \bar{y}) = \frac{1}{K} \sigma^2.
	\end{equation}
	Now, compute \(\operatorname{Var}(\hat{y}_k - \bar{y})\):
	\begin{align}
		\operatorname{Var}(\hat{y}_k - \bar{y}) &= \sigma^2 + \frac{\sigma^2}{K} - 2 \left( \frac{\sigma^2}{K} \right) \\
		&= \sigma^2 \left( 1 + \frac{1}{K} - \frac{2}{K} \right) \\
		&= \sigma^2 \left(1 - \frac{1}{K}\right).
	\end{align}
	
	Thus, 
	\begin{equation}
		\mathbb{E}[L_{\text{var}}] = \frac{1}{K} \cdot K \cdot \sigma^2 \left(1 - \frac{1}{K}\right) = \sigma^2 \left(1 - \frac{1}{K}\right).
	\end{equation}
	Simplify:
	\begin{equation}
		\mathbb{E}[L_{\text{var}}] = \sigma^2 \left( \frac{K - 1}{K} \right).
	\end{equation}
	
	The expected regression loss is:
	\begin{equation}
		\mathbb{E}[L_{\text{reg}}] = \mathbb{E}\left[(\hat{y} - y)^2\right] = \mathbb{E}\left[(\bar{y} - y)^2\right].
	\end{equation}
	Since we can obtain the prediction \(\hat{y}\) for the test sample \(\boldsymbol{x}_{\text{test}}\) with the average over all augmented versions:
	\begin{equation}
		\hat{y} = \bar{y} = \frac{1}{K}\sum_{k=1}^K \hat{y}_k.
	\end{equation}
	Under Assumption 1 (Unbiased Augmentation):
	\begin{equation}
		\mathbb{E}_{\mathcal{T}_{\text{aug}}}[\hat{y}_k] = y, \quad \forall k.
	\end{equation}
	the expected value of \(\bar{y}\) is equal to \(y\),
	\begin{equation}
		\mathbb{E}[\bar{y}] = \mathbb{E}\left[\frac{1}{K}\sum_{k=1}^K \hat{y}_k\right] = \frac{1}{K}\sum_{k=1}^K \mathbb{E}[\hat{y}_k] = y.
	\end{equation}
	Therefore, we have:
	\begin{equation}
		\mathbb{E}[L_{\text{reg}}] = \mathbb{E}\left[(\bar{y} - \mathbb{E}[\bar{y}])^2\right] = \operatorname{Var}(\bar{y}) = \frac{\sigma^2}{K}.
	\end{equation}
	
	So far, we have:
	\begin{align}
		\mathbb{E}[L_{\text{reg}}] &= \frac{\sigma^2}{K}, \label{equ:29}\\
		\mathbb{E}[L_{\text{var}}] &= \sigma^2 \left( \frac{K - 1}{K} \right). \label{equ:30}
	\end{align}
	Substitute Equation \ref{equ:29} into Equation \ref{equ:30}:
	\begin{equation}
		\mathbb{E}[L_{\text{reg}}] = \frac{\sigma^2}{K} = \frac{1}{K} \cdot \mathbb{E}[L_{\text{var}}] \left( \frac{K}{K - 1} \right) = \frac{\mathbb{E}[L_{\text{var}}]}{K - 1}.
	\end{equation}
	Recall that for \(K \geq 2\), the following inequality holds:
	\begin{equation}
		\frac{1}{K - 1} \leq \frac{2}{K}.
	\end{equation}
	
	\textbf{Therefore, we have:}
	\begin{equation}
		\mathbb{E}[L_{\text{reg}}] = \frac{\mathbb{E}[L_{\text{var}}]}{K - 1} \leq \frac{2\,\mathbb{E}[L_{\text{var}}]}{K}.
	\end{equation}
	
	\hfill $\square$
	
	Under Assumptions 1 and 2, we have established that minimizing the variance loss \(\mathbb{E}[L_{\text{var}}]\) effectively reduces the expected regression error \(\mathbb{E}[L_{\text{reg}}]\), with the bound improving as the number of augmentations \(K\) increases. This theoretical result justifies the effectiveness of our variance minimization strategy in test-time training.

	\section{Algorithm}
	\label{appenx_sec_algo}
	As shown in Algorithm \ref{alg:ttt_qpn}, our approach consists of two phases. During training, we first extract features through an encoder and decompose them into periodic and aperiodic components. The periodic component is modeled using sinusoidal functions with learned parameters (frequency, phase, bias, and velocity), while the aperiodic component is captured through a continuous-time framework using cubic spline interpolation. During test-time adaptation, we generate $K$ augmented views of each test sample and employ a differential learning rate strategy: applying smaller learning rates to periodic components' batch normalization parameters to maintain stable cardiac patterns, while using larger learning rates for aperiodic components to enable flexible patient-specific adaptation.
	
	\begin{algorithm}[t]
		\caption{Test-time Training for Quasi-Period Network}
		\label{alg:ttt_qpn}
		\begin{algorithmic}[1]
			\Require Training data $\mathcal{D}_{train} = \{(\boldsymbol{x}_i, y_i)\}_{i=1}^N$, test sample $\boldsymbol{x}_{test}$
			\Ensure Adapted model parameters $\theta$
			
			\State // Training Phase
			\For{each training iteration}
			\State $\boldsymbol{z} \gets \text{Enc}(\boldsymbol{x})$ \Comment{Initial feature extraction}
			
			\State // Periodic Component
			\State $f, \phi, b, v \gets p(\boldsymbol{z})$ \Comment{Extract periodic parameters}
			\State $\boldsymbol{\hat{z}}^{period} \gets \cos(2\pi(ft-\phi)) + \sin(2\pi(ft-\phi)) + vt + b$
			
			\State // Aperiodic Component
			\State $\boldsymbol{z}' \gets \boldsymbol{z} - \boldsymbol{\hat{z}}^{period}$ \Comment{Residual features}
			\State $\boldsymbol{V}(t) \gets \text{CubicSpline}(\boldsymbol{z}'_t)$ \Comment{Continuous path}
			\State $\boldsymbol{\hat{z}}^{aperiod} \gets \boldsymbol{z}'_0 + \int_0^T f_{\theta}(\boldsymbol{\hat{z}}^{aperiod}_t, t) d\boldsymbol{V}(t)$
			
			\State // Loss Computation
			\State Calculate training loss $\mathcal{L}_{total}$ using Eq. \ref{eq:train_total}
			\State Update model parameters
			\EndFor
			
			\State // Test-time Training Phase
			\For{each test sample $\boldsymbol{x}_{test}$}
			\State // Generate Augmented Samples
			\State $\{\boldsymbol{x}_{test}^k\}_{k=1}^K \gets \mathcal{T}_{aug}(\boldsymbol{x}_{test})$
			
			\For{each adaptation iteration}
			\For{$k = 1$ to $K$}
			\State Forward pass $\boldsymbol{x}_{test}^k$ through network
			\State Compute predictions $\hat{y}_k$
			\EndFor
			
			\State // Compute Test-time Losses
			Calculate test-time loss $\mathcal{L}_{test}$ using Eq. \ref{eq:loss_test}
			
			\State // Differential Adaptation
			\State Update periodic BN parameters with small learning rate
			\State Update aperiodic BN parameters with large learning rate
			\EndFor
			\EndFor
		\end{algorithmic}
	\end{algorithm}
	
	\section{Experiment}
	\label{append_sec_exp}
	\subsection{Evaluation Metrics}
	\label{sec:append_metrics}
	MAE measures the average absolute differences between predicted and ground truth LVEF values. RMSE emphasizes larger prediction errors by computing the square root of the mean squared differences. MAPE calculates the percentage error relative to the ground truth value, providing a scale-independent assessment of model performance. These metrics are formally defined as:
	
	\begin{equation}
		\text{MAE} = \frac{1}{n}\sum_{i=1}^n |y_i - \hat{y}_i|,
	\end{equation}
	
	\begin{equation}
		\text{RMSE} = \sqrt{\frac{1}{n}\sum_{i=1}^n (y_i - \hat{y}_i)^2},
	\end{equation}
	
	\begin{equation}
		\text{MAPE} = \frac{100\%}{n}\sum_{i=1}^n |\frac{y_i - \hat{y}_i}{y_i}|,
	\end{equation}
	where $y_i$ and $\hat{y}_i$ denote the ground truth and predicted LVEF values respectively, and $n$ is the number of test samples.
	
	\begin{table*}[t]
	\centering
	\setlength{\tabcolsep}{4pt}  
	\newcolumntype{P}[1]{>{\centering\arraybackslash}p{#1}} 
	\caption{\textbf{All Ablation Study Results of Key Components.} Analysis of three key components: QP-Net (Quasi-Period Network), LR (Learning Rate Strategy), and VM (Variance Minimization).}
	\resizebox{\linewidth}{!}{
		\begin{tabular}{P{1.1cm}P{1.1cm}P{1.1cm}|P{1.5cm}P{1.5cm}P{1.5cm}|P{1.5cm}P{1.5cm}P{1.5cm}|P{1.5cm}P{1.5cm}P{1.5cm}}
			\toprule
			QP-Net &  LR & VM  & \multicolumn{3}{c|}{Pre-School} & \multicolumn{3}{c|}{School Age} & \multicolumn{3}{c}{Adolescence} \\
			&   &  & MAE & MAPE & RMSE & MAE & MAPE & RMSE & MAE & MAPE & RMSE \\
			\midrule
			& & 
			& 8.129 & 0.1681 & 10.09 & 8.267 & 0.1454 & 9.999 & 8.222 & 0.1498 & 10.2 \\
			\rowcolor[gray]{0.9} \cmark & & 
			& 7.949 & 0.1666 & 9.909 & 7.400 & 0.1368 & 8.959 & 7.524 & 0.1451 & 9.488 \\
			\cmark & \cmark & 
			& 7.909 & 0.1656 & 9.878 & 7.480 & 0.1331 & 9.031 & 7.501 & 0.1403 & 9.383 \\
			\rowcolor[gray]{0.9} & & \cmark
			& 7.842 & 0.1640 & 9.714 & 6.683 & 0.1260 & 8.767 & 7.146 & 0.1390 & 9.014 \\
			\cmark &  & \cmark
			& 7.283 & 0.1619 & 9.307 & 6.708 & 0.1243 & 8.437 & 6.988 & 0.1344 & 9.002 \\
			\rowcolor[gray]{0.9} \cmark & \cmark & \cmark 
			& 7.235 & 0.1611 & 9.290 & 6.706 & 0.1244 & 8.432 & 6.980 & 0.1344 & 8.950 \\
			\bottomrule
		\end{tabular}
	}
	\label{tab:ablation_full}
\end{table*}
	
	As for AUROC, we follow the clinical guidelines and set four critical LVEF thresholds: 35\%, 40\%, 45\%, and 50\%. These thresholds are clinically significant as they correspond to different levels of cardiac dysfunction. For each threshold, we compute the AUROC score by treating LVEF prediction as a binary classification problem, where values below the threshold indicate potential cardiac dysfunction. The mean AUROC across all thresholds provides a comprehensive assessment of the model's ability to identify clinically relevant cardiac conditions.

	\subsubsection{Baseline Implementation}
	Segmentation-based methods follow a two-step process: first segmenting the left ventricle in each frame of the echocardiogram video, then calculating LVEF based on the end-diastolic volume (EDV) and end-systolic volume (ESV). Specifically, after obtaining segmentation masks for all frames, the frames with maximum and minimum left ventricular volumes are identified as end-diastolic and end-systolic frames, respectively. LVEF is then calculated using the following formula:
	\begin{equation}
		\text{LVEF} = \frac{\text{EDV} - \text{ESV}}{\text{EDV}} \times 100\%,
	\end{equation}
	where EDV and ESV are computed from the segmentation masks using standard clinical volume estimation methods.
	
	Vision-language models approach LVEF prediction as a cross-modal similarity task. These methods first encode the echocardiogram video into visual tokens through a vision encoder. Simultaneously, they construct a series of language tokens representing different LVEF values (e.g., "The left ventricular ejection fraction is X percent", where X ranges from 0 to 100). The predicted LVEF is determined by finding the language token that exhibits the highest similarity score with the visual tokens in the joint embedding space.
	
	\subsubsection{All results for Table \ref{tab:ablation} }
	
	We show the all results from three cohorts in Table \ref{tab:ablation_full}.

	
\end{document}